\documentclass[review,10pt,a4paper]{elsarticle}
\graphicspath{{./figures/}}
\usepackage{amsmath,amssymb,amsfonts} 
\usepackage{graphicx}                
\usepackage{float}                
\usepackage{multirow}                    
\usepackage{booktabs}                     
\usepackage{caption}    
\usepackage[utf8]{inputenc}
\usepackage[labelfont=bf, format=plain]{caption} 
\usepackage{longtable,tabularx}    
\usepackage{subcaption}                  
\usepackage{xcolor}                    
\usepackage{geometry}                
\usepackage[colorlinks=true, allcolors=blue]{hyperref} 
\usepackage{tcolorbox} 
\tcbset{colback=gray!5!white, colframe=gray!75!black, fonttitle=\bfseries, title=The pseudo-code of training friction coefficient model}
\restylefloat{figure} 
\restylefloat{table}
\biboptions{authoryear}

\begin{document}
\begin{frontmatter}
\title{OnSiteVRU: A High-Resolution Trajectory Dataset for High-Density Vulnerable Road Users}

\author[label1,label2]{Zhangcun Yan}
\author[label1,label3]{Jianqiang Li}
\author[label1]{Peng Hang}
\author[label1,cor1]{Jian Sun}
\ead{sunjian@tongji.edu.cn}
\cortext[label1]{Corresponding author.}

\address[label2]{School of Traffic and Transportation Engineering, Changsha University of Science \& Technology, No. 960, 2nd section of Wanjiali Road, Changsha, Hunan Province, 410114, China}
\address[label1]{Key Laboratory of Road and Traffic Engineering of Ministry of Education \& School of Transportation Engineering, Tongji University. No.4800 Cao’an Road, Shanghai, 201804, China.}
\address[label3]{ School of Automobile and Transportation, Xihua University Chengdu Sichuan, China}

\begin{abstract}
With the acceleration of urbanization and the growth of transportation demands, the safety of vulnerable road users (VRUs, such as pedestrians and cyclists) in mixed traffic flows has become increasingly prominent, necessitating high-precision and diverse trajectory data to support the development and optimization of autonomous driving systems. However, existing datasets fall short of capturing the diversity and dynamics of VRU behaviors, making it difficult to meet the research demands of complex traffic environments. To address this gap, this study developed the OnSiteVRU datasets, which cover a variety of scenarios, including intersections, road segments, and urban villages. These datasets provide trajectory data for motor vehicles, electric bicycles, and human-powered bicycles, totaling approximately 17,429 trajectories with a precision of 0.04 seconds. The datasets integrate both aerial-view natural driving data and onboard real-time dynamic detection data, along with environmental information such as traffic signals, obstacles, and real-time maps, enabling a comprehensive reconstruction of interaction events. The results demonstrate that VRU\_Data outperforms traditional datasets in terms of VRU density and scene coverage, offering a more comprehensive representation of VRU behavioral characteristics. This provides critical support for traffic flow modeling, trajectory prediction, and autonomous driving virtual testing. The dataset is publicly available for download at: ~\url{https://www.kaggle.com/datasets/zcyan2/mixed-traffic-trajectory-dataset-in-from-shanghai}.

\end{abstract}

\end{frontmatter}

\section{Introduction}

With the acceleration of urbanization and the increasing demand for transportation, the complexity of mixed traffic flows (including conventional vehicles, autonomous vehicles, pedestrians, cyclists, etc.) has grown significantly. Safety concerns for Vulnerable Road Users (VRUs, such as pedestrians and cyclists) have become particularly prominent. Autonomous vehicles must rapidly and accurately identify and respond to VRU behaviors in complex traffic environments. However, existing testing methods and datasets often fail to comprehensively capture the diversity and dynamism of VRU behaviors. While several high-precision trajectory datasets have been developed for training and validating autonomous driving algorithms, these datasets still exhibit shortcomings in VRU-related scenario coverage, data quality, and diversity. For example, existing datasets may lack sufficient representation of VRU behavior patterns in complex intersections or sudden traffic incidents, making it challenging for autonomous systems to effectively ensure VRU safety in real-world applications. Consequently, developing a more representative, diverse, and high-precision VRU trajectory dataset is critically important. Yet, this task faces numerous challenges, including the high cost of VRU data collection, the complexity of behavioral annotation, and ensuring the dataset’s broad applicability to VRU behaviors.  

The distribution characteristics of mixed urban traffic flows—particularly VRU behavior patterns—in scenarios such as intersections, mixed-use road sections (shared by motorized and non-motorized traffic), and urban villages exhibit high complexity and dynamism. Compared to traffic flows dominated by motor vehicles, these environments feature greater unpredictability, such as the randomness of VRU behaviors, diverse path choices, and frequent interactions with vehicles. For instance, at intersections, VRUs often exhibit sudden turns or stops; in mixed-use road sections, collision risks between motorized and non-motorized users increase significantly; and in narrow urban village areas, the intermingling of VRUs and vehicles further intensifies traffic complexity. These features make it extremely difficult for computer vision algorithms to extract complete trajectories due to occlusion-induced partial data loss and severe noise issues. Even state-of-the-art algorithms struggle to collect high-precision traffic flow trajectory data. Additionally, these characteristics lead to uncertain vehicle paths, unstable traffic flows, and highly heterogeneous driving behaviors, significantly raising traffic conflict rates and collision risks, particularly endangering VRU safety.  

High-resolution trajectory data provides comprehensive and accurate microscopic behavioral information, especially for detailed characterization of VRU behaviors. This data is not only a critical resource for traffic flow theory research but also forms a foundational pillar for the digital transformation of smart cities and intelligent transportation systems. As a vital carrier of microscopic traffic behavior information, high-resolution trajectory data plays an irreplaceable role in modeling and analyzing motorized-non-motorized interaction risks in shared intersection spaces. However, existing datasets still lack sufficient coverage, quality, and diversity in VRU-related scenarios. For example, they often fail to adequately capture VRU behaviors in extreme weather, complex intersections, or sudden traffic events, limiting the ability of autonomous systems to ensure VRU safety in practice. To clarify research progress on mixed traffic flow trajectories in shared intersection spaces—particularly studies focused on VRU behaviors—this paper systematically reviews existing work from two perspectives: trajectory datasets for shared intersection spaces and algorithms for improving trajectory precision. Meanwhile, the complex dynamic characteristics of shared intersection spaces mean that motorized-non-motorized interactions are influenced by coupled factors, exhibiting features of complex dynamic systems. This makes it challenging to decouple the mechanisms underlying interaction risk emergence and development from a macro perspective and to model the risk evolution of interactions at a micro level.  

Currently, significant issues persist in the collection, processing, and application of VRU data. First, existing research primarily relies on high-resolution sensors (e.g., LiDAR, cameras) and multi-source data fusion techniques to capture VRU microscopic behaviors, aiming to enhance data precision and coverage. For instance, some studies employ deep learning algorithms to predict and reconstruct VRU trajectories to address occlusion and noise issues. Others use simulation environments or crowdsourced data to supplement VRU behavior data that is difficult to obtain in real-world scenarios. However, these approaches have notable limitations: (1) High data collection costs, especially in complex scenarios (e.g., extreme weather, nighttime, or high-density traffic areas); (2) Insufficient diversity and representativeness of VRU behaviors in existing datasets, with inadequate coverage of special groups (e.g., children, elderly) or atypical behaviors (e.g., sudden running, jaywalking); (3) Privacy concerns that restrict data openness and sharing, limiting dataset scale and applicability; and (4) The complexity and subjectivity of data annotation, which may introduce errors affecting model training. While technical progress has been made, further improvements in data quality, diversity, and privacy protection are needed to better support VRU safety assurance and mixed traffic flow optimization.  

To address these challenges, this study collects VRU trajectory data from typical Chinese traffic scenarios (e.g., intersections, road sections, urban villages) and integrates multi-source sensing technologies such as computer vision algorithms and high-precision radar to construct a high-quality, diverse, and open-source high-precision VRU trajectory dataset. This dataset covers complex intersections, mixed-use road sections, urban villages, and diverse conditions such as extreme weather, nighttime environments, and sudden traffic events, comprehensively reflecting the diversity and dynamism of VRU behaviors. Additionally, advanced anonymization techniques and automated annotation tools are employed to ensure data privacy while improving annotation efficiency and accuracy. By open-sourcing this dataset, we aim to provide academia and industry with a critical resource to advance VRU behavior modeling, autonomous driving algorithm optimization, and traffic safety research, thereby supporting the digital transformation of smart cities and intelligent transportation systems.

\section{Literature review}

The construction of VRU (Vulnerable Road User) trajectory datasets is a foundational challenge for autonomous driving safety research, requiring a balance between technological innovation and scenario adaptability. With accelerating urbanization and rapid advancements in autonomous driving technology, the safety of VRUs---such as pedestrians and cyclists---has garnered increasing attention. Autonomous systems must accurately identify and respond to VRU behaviors in complex traffic environments, where high-precision, diverse VRU trajectory datasets are critical for algorithm optimization and safety validation. However, existing datasets exhibit significant shortcomings in scenario coverage, data quality, and diversity, failing to meet the demands of research in intricate traffic environments. A review of publicly available datasets is provided in Table~\ref{tab:datasets_table}.

Among internationally recognized datasets, NGSIM has been widely used for vehicle trajectory analysis, but concerns remain regarding data accuracy and its limited support for VRU-oriented studies \citep{punzo2011assessment,coifman2017critical}. Similarly, HighD focuses on naturalistic vehicle trajectories on German highways and is therefore less suitable for studying dense motorized--non-motorized interactions in urban environments \citep{krajewski2018highd}. At intersections, Ko-PER and LUMPI provide valuable observations for multi-participant traffic analysis, while InD captures naturalistic road-user trajectories at German intersections using drones; however, these datasets are still limited in scene diversity and in representing complex mixed-traffic conditions \citep{strigel2014ko,busch2022lumpi,bock2020ind}.

Asian datasets exhibit stronger regional characteristics. ApolloScape provides urban-road data for autonomous driving, SinD offers drone-based observations at signalized intersections in China, and V2X-Seq supports vehicle--infrastructure cooperative perception and forecasting \citep{huang2018apolloscape,xu2022drone,yu2023v2x}. In India, IDD-X emphasizes dense and unstructured traffic scenes, which is valuable for understanding region-specific traffic complexity \citep{parikh2024idd}. In addition, large-scale urban traffic monitoring efforts such as pNEUMA and top-view trajectory datasets further enrich the available data resources, but they still do not fully cover high-density VRU interactions across intersections, road segments, and urban-village scenarios \citep{barmpounakis2020new,yang2019top,lerner2007crowds}.

The limitations of existing datasets stem from multifaceted technical challenges in VRU trajectory collection and processing. First, the dynamic and stochastic nature of VRU behaviors in complex scenarios complicates data acquisition. For example, in shared intersection spaces, dense interactions between VRUs and vehicles lead to sensor occlusion, causing trajectory noise or partial data loss even with high-precision radar and computer vision algorithms. Second, data annotation complexity and high costs persist. Manual annotation of behaviors like sudden running or jaywalking risks subjective errors, while automated tools often struggle with generalization in complex scenarios.

Recent technical advancements aim to enhance data quality. For example, realistic behavior extraction frameworks can better align observed road-user behavior with safety-related modeling assumptions \citep{certad2023extraction}. Trajectory reconstruction methods integrating social force models and particle filtering have also been used to improve the resolution and continuity of non-motorized trajectories in shared spaces \citep{yan2023high}. Meanwhile, high-resolution trajectory data have supported fine-grained modeling of motorized--non-motorized interaction risk in intersection center areas \citep{yan2025investigating}. Nevertheless, open datasets that simultaneously provide high-density VRU trajectories, diverse Chinese urban scenarios, and rich contextual information such as signal states and HD maps remain limited.

In summary, the development of VRU trajectory datasets demands both technological breakthroughs and scenario-specific adaptations. Our research team addresses this gap by constructing an open-source VRU trajectory dataset for China's complex traffic environments. Leveraging multimodal data from intersections, road sections, and urban villages, combined with high-precision radar and computer vision algorithms, the dataset emphasizes high-density interaction events. By systematically reviewing VRU dataset advancements and technical challenges, this study provides theoretical and practical guidance for future dataset development. As technology and ecosystems evolve, VRU datasets will play an increasingly vital role in autonomous driving safety research, laying a robust foundation for global transportation safety and intelligent traffic systems.

\begin{table}[htbp]
\centering
\caption{Summary of Dataset Information}
\label{tab:datasets_table}
\resizebox{\textwidth}{!}{
\begin{tabular}{llllll}
    \toprule
    \textbf{Dataset Name} & \textbf{Location} & \textbf{Setting} & \textbf{Composition} & \textbf{Collection Method} & \textbf{Purpose} \\
    \midrule
    Mix-t                  & India       & Urban roads        & Mixed traffic flow & Roadside camera         & Mixed flow modeling \\
    BIWI Hotel             & Switzerland & Entrance/Exit      & Pedestrians        & Roadside camera         & Behavior modeling \\
    BIWI ETH               & Switzerland & Campus roads       & Pedestrians        & Roadside camera         & Behavior modeling \\
    Crowds~\cite{lerner2007crowds} & USA         & Campus and roads   & Pedestrians        & Roadside camera         & Behavior modeling \\
    Ko-PER~\cite{strigel2014ko}    & Germany     & Urban intersections & Mixed traffic flow & Roadside camera         & Behavior modeling \\
    VRU                    & Germany     & Urban intersections & Mixed traffic flow & Roadside camera         & Behavior modeling \\
    DUT / CITR             & Germany     & Campus roads       & Mixed traffic flow & Drone                   & Behavior modeling \\
    Stanford Drone         & USA         & Campus             & Mixed traffic flow & Drone                   & Behavior modeling \\
    LUMPI~\cite{busch2022lumpi}    & Germany     & Intersections      & Mixed traffic flow & Radar, roadside camera  & Traffic safety analysis \\
    UniD                   & Germany     & Campus roads       & Mixed traffic flow & Drone                   & Behavior prediction \\
    INTERACTION            & USA/China   & Intersections      & Vehicles           & Drone, roadside         & Behavior modeling \\
    V2X-Seq~\cite{yu2023v2x}       & China       & Intersections      & Mixed traffic flow & Radar, roadside camera  & Driving testing \\
    InD~\cite{bock2020ind}         & Germany     & Intersections      & Mixed traffic flow & Drone                   & Interaction behavior \\
    ApolloScape~\cite{huang2018apolloscape} & China & Urban roads    & Mixed traffic flow & Radar, roadside camera  & Autonomous driving \\
    TRAF                   & India       & Intersections      & Mixed traffic flow & Drone                   & Behavior modeling \\
    Waymo                  & USA         & Urban roads        & Mixed traffic flow & Onboard radar           & Driving testing \\
    IDD-X~\cite{parikh2024idd}     & India       & Intersections      & Mixed traffic flow & Onboard camera          & Behavior modeling \\
    HighD~\cite{krajewski2018highd} & Germany    & Highways           & Vehicles           & Drone                   & Driving testing \\
    SinD~\cite{xu2022drone}        & China       & Intersections      & Mixed traffic flow & Drone                   & Driving testing \\
    \bottomrule
\end{tabular}}
\end{table}

\section{Dataset Construction}
In this section, we outline the detailed , covering location selection, data collection, data processing, map generation, and other key aspects.

\subsection{Data Collection Location Selection}

The core objective of the VRU (Vulnerable Road Users) dataset is to provide data support for the development of key technologies in complex driving environments, with a focus on optimizing driving safety control algorithms, designing active traffic safety management systems, and training trajectory prediction models for non-motorized vehicles. To achieve these objectives, the selection of data collection locations must follow specific principles to ensure data validity and scene representativeness.

First, priority should be given to high-accident areas, particularly locations with high VRU accident rates according to traffic accident statistics. For example, intersections—characterized by multiple conflict points and large blind spots—are ideal for capturing high-risk behaviors such as sudden braking and aggressive maneuvers, providing valuable data for quantitative safety risk analysis. Unstructured road segments (e.g., urban villages and construction zones) with ambiguous traffic rules and frequent VRU-motor vehicle interactions can reflect complex traffic behaviors. Similarly, mixed-traffic roads without signal control exhibit highly dynamic traffic flows, making them suitable for studying conflict behaviors in the absence of regulatory constraints.

Additionally, areas with complex traffic flows should be key targets for data collection. This includes mixed-phase intersections (e.g., unprotected left-turn phases and shared release phases for motorized and non-motorized traffic), which capture dynamic interactions between vehicles and VRUs. Multi-phase signalized intersections (e.g., two-phase or three-phase signal control) are useful for studying the impact of traffic signals on traffic flow, while unsignalized intersections reveal self-organizing traffic behaviors in the absence of signal constraints.

Second, to ensure broad applicability, the selected locations should cover diverse road types and VRU groups. The dataset should include urban arterial roads, secondary roads, local streets, and alleyways to reflect traffic characteristics across different road classifications. VRU classifications should distinguish among pedestrians, human-powered bicycles, and electric bicycles, with special attention to the unique trajectory patterns of vulnerable road users to enhance dataset representativeness.

Finally, to ensure data quality, the selection of data collection locations should minimize external interference. In terms of environmental isolation, preference should be given to intersections without external traffic disruptions, such as parking lots or bus stations, to avoid contamination from non-target traffic flows. Regarding equipment deployment, cameras and other sensors should be installed at elevated positions to ensure complete coverage of the target area. Additionally, the data collection process must not occupy traffic lanes or obstruct normal traffic flow, as this could artificially alter driver behavior.

By systematically applying these principles, the dataset can maintain high data quality while providing valuable scenario samples for autonomous driving algorithm validation and traffic policy formulation, promoting the widespread application of the VRU dataset in traffic safety research.

\subsection{Data collection}
\begin{figure}[htbp]
\tiny{
    \centering
    \begin{subfigure}[b]{0.24\textwidth}
        \centering
        \includegraphics[width=\textwidth]{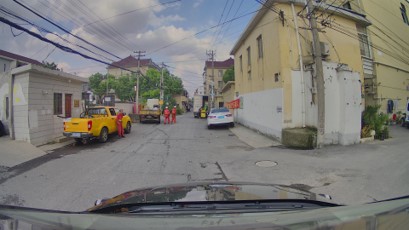}
        \caption{Blurred road boundaries}
    \end{subfigure}
    \begin{subfigure}[b]{0.24\textwidth}
        \centering
        \includegraphics[width=\textwidth]{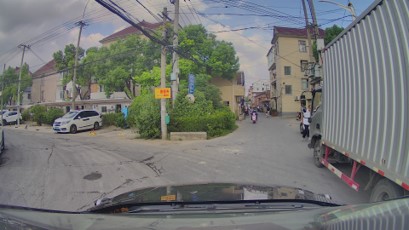}
        \caption{Complex intersections}
    \end{subfigure}
    \begin{subfigure}[b]{0.24\textwidth}
        \centering
        \includegraphics[width=\textwidth]{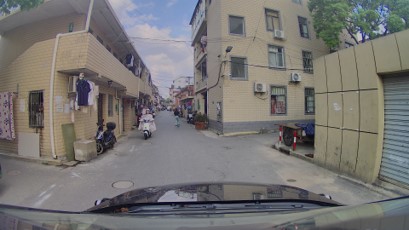}
        \caption{VRU-dense roads}
    \end{subfigure}
    \begin{subfigure}[b]{0.24\textwidth}
        \centering
        \includegraphics[width=\textwidth]{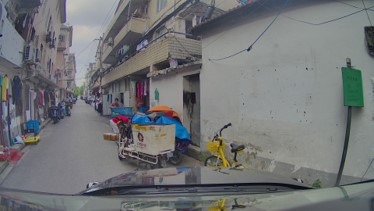}
        \caption{Unexpected pedestrian}
    \end{subfigure}
    \vspace{0.3cm} 
    \begin{subfigure}[b]{0.24\textwidth}
        \centering
        \includegraphics[width=\textwidth]{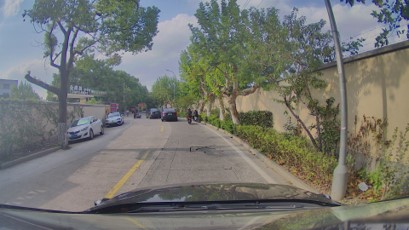}
        \caption{Heterogeneous traffic flow}
    \end{subfigure}
    \begin{subfigure}[b]{0.24\textwidth}
        \centering
        \includegraphics[width=\textwidth]{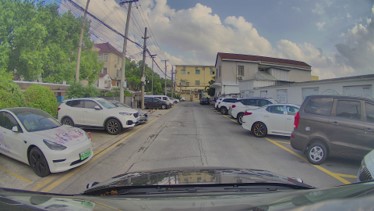}
        \caption{Roadside parking obstacles}
    \end{subfigure}
    \begin{subfigure}[b]{0.24\textwidth}
        \centering
        \includegraphics[width=\textwidth]{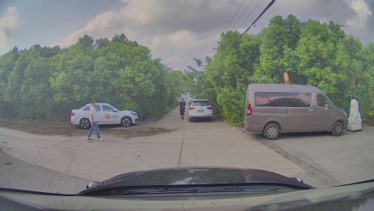}
        \caption{Multi-VRU interactions}
    \end{subfigure}
    \begin{subfigure}[b]{0.24\textwidth}
        \centering
        \includegraphics[width=\textwidth]{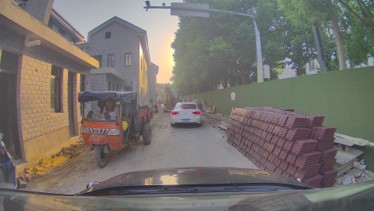}
        \caption{Emergency meeting}
    \end{subfigure}
    \caption{Complex scenarios in urban villages}
        \label{fig: Wangjianong Urban Village}}
\end{figure}

Based on the above principles, 20 potential data collection sites were initially identified using satellite maps. Following multiple field surveys, four intersections, two road segments, and one urban village were selected as the final study locations.
\begin{itemize}
    \item Longchang Road - Changyang Road Intersection: This intersection employs a three-phase signal control scheme. The north-south left-turn and through movements share one phase, the east-west left-turns share another phase, and the east-west through movements share the third phase. During each phase, motorized and non-motorized vehicles have equal priority, while right turns are unprotected.

    \item Ningwu Road - Hejian Road Intersection: This intersection utilizes a two-phase signal control scheme. The east-west left-turn and through movements share one phase, while the north-south left-turn and through movements share another phase. Mixed traffic is permitted at this intersection, allowing different types of vehicles to travel together.

    \item Jianhe Road - Xianxia Road Intersection: This intersection operates under a two-phase signal control scheme. It experiences a high volume of non-motorized traffic in the east-west direction, leading to significant conflicts between left-turn traffic flows (east-to-south and west-to-north) and through non-motorized traffic.

    \item Moyu Road - Changji East Road Intersection: Located in Anting Town, Shanghai, this intersection connects Moyu Road, a major arterial road carrying the primary traffic flow, and Changji East Road, a secondary road with lower traffic volume but high non-motorized vehicle flow. The intersection uses a three-phase signal control scheme, featuring a protected left-turn phase for the north-south direction on Moyu Road, while the east-west left-turn, through, and right-turn movements proceed simultaneously.

    \item Caoyang Road: Situated in Putuo District, Shanghai, this road segment features soft isolation and lane markings separating motorized and non-motorized lanes. Traffic volume is moderate, with frequent interactions between motorized and non-motorized vehicles. The selected segment is located 40 meters away from an intersection to avoid signal interference. It experiences frequent violations by non-motorized vehicles occupying motorized lanes, creating mixed traffic conditions. The road provides an unobstructed elevated view, making it suitable for high-position traffic flow data collection.

    \item Anyuan Road: Located in Jing’an District, Shanghai, this road is classified as a local street with a single-slab cross-section. Opposing motorized lanes are separated by lane markings, while same-direction motorized and non-motorized lanes are divided by soft isolation. Traffic volume is high, particularly for electric bicycles and bicycles, resulting in frequent high-density interactions between motorized and non-motorized vehicles.

    \item Wangjianong Urban Village: Located in Minhang District, Shanghai, this area consists of low-grade, primarily unstructured roadways, as shown in ~\ref{fig: Wangjianong Urban Village}.  The roads frequently experience interference from electric bicycles, bicycles, pedestrians, and obstacles, affecting vehicle movement. The area includes various complex scenarios, such as mixed motorized and non-motorized traffic, narrow roads, unclear road boundaries, and intricate intersections with obstacles, making it a typical representation of urban village roads in Chinese cities.
\end{itemize}

\section{Data Processing Procedure}
The data collection for intersections and road segments was conducted using a roadside high-altitude overhead recording method to capture operational video data. According to the Shanghai UAV Flight Guide, drone flights are prohibited within 50 meters of major roads. Additionally, drone-based video collection suffers from vibration interference, which significantly impacts the detection and trajectory tracking of small objects such as non-motorized vehicles. To obtain more stable video footage, cameras were installed on the rooftops of adjacent buildings to record intersection operations.

To ensure the collection of a sufficient number of interaction event samples, the filming time was selected based on peak non-motorized traffic hours in Shanghai—namely, the morning peak (7:00–9:00) and evening peak (17:00–19:00). Using 4K long-lens cameras, a total of six hours of video data was collected from four intersections, with one hour of video data recorded for each of the two road segments. The video footage is clear and captures the complete movement of vehicles from the stop line at the intersection to the crosswalk at the exit lane.

To calibrate the cameras and convert pixel coordinates into world coordinates, intersection geometric parameters were measured, and signal control schemes were recorded.

For data collection in urban villages, an experimental vehicle was equipped with various perception modules, including LiDAR, millimeter-wave radar, integrated navigation systems, industrial control computers, and smart cameras. While driving through the area, the vehicle continuously collected real-time scene environment videos from the front, left, right, and rear perspectives, detecting surrounding objects. LiDAR was used to obtain point cloud data, while the onboard IMU and GPS provided additional vehicle state information.

\subsection{Vehicle Detection and Trajectory Tracking}
\begin{figure}
    \centering
    \includegraphics[width=1\linewidth]{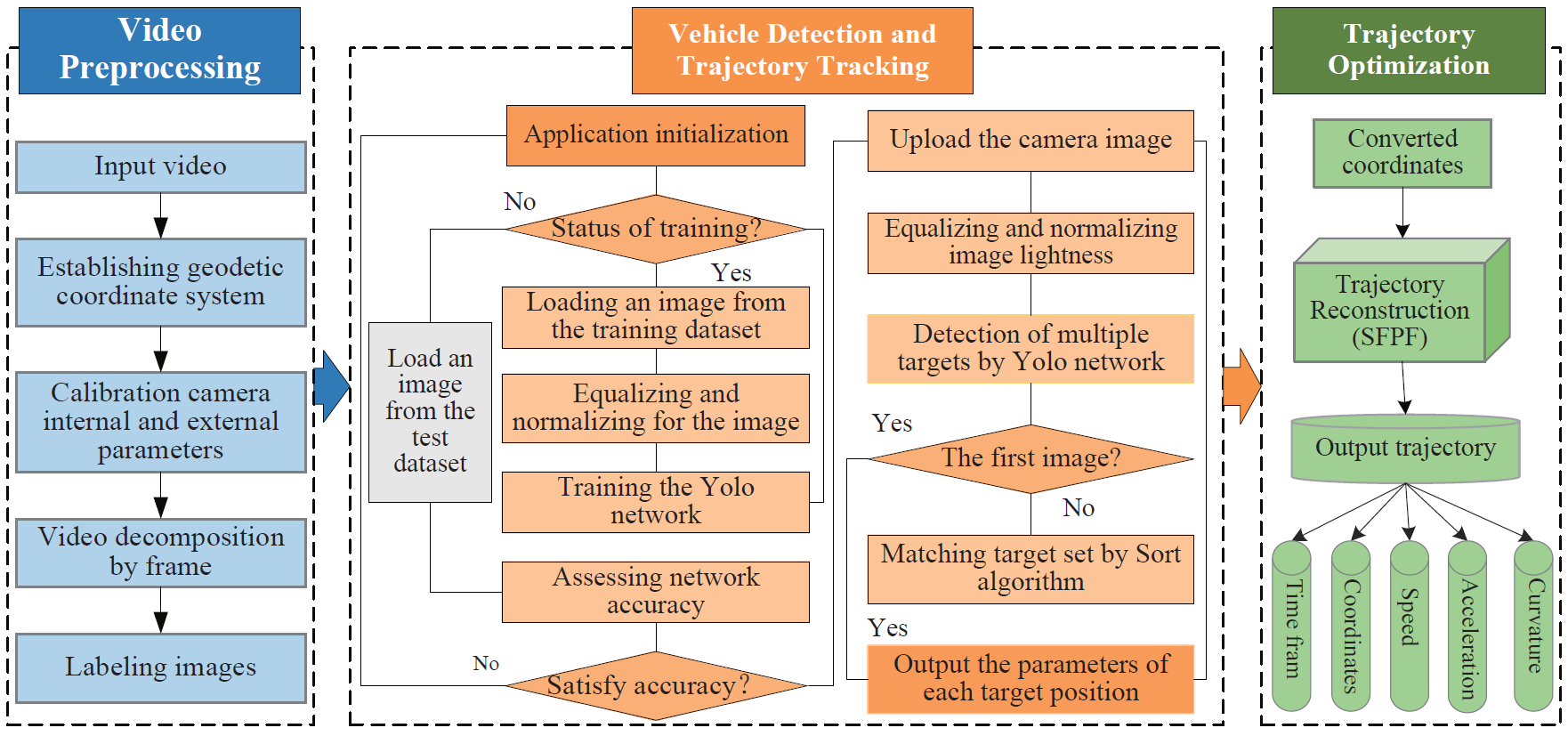}
    \caption{Vehicle Detection and Tracking Framework}
    \label{fig:flowchart}
\end{figure}

This study adopts the technical approaches of manual trajectory extraction and semi-automated trajectory extraction software, integrating open-source software modules to develop an automated trajectory extraction platform. The platform incorporates the YOLOv7 object detection algorithm, the DeepSORT multi-object tracking algorithm, and camera calibration and coordinate transformation functions from OpenCV~\cite{yan2025investigating}. By combining these computer vision algorithms, a vehicle trajectory extraction platform is constructed. The specific architecture of the platform is shown in Figure~\ref{fig:flowchart}.

The trajectory extraction platform consists of three main functional modules:
\begin{itemize}
    \item \textbf{Video Preprocessing Module:} This module performs camera intrinsic and extrinsic parameter calibration, extracts frames from videos, and annotates detection targets.
    \item \textbf{Object Detection and Tracking Module:} Using the annotated dataset, a YOLOv7 object detection network is trained. The trained model then processes video frames sequentially, detecting different types of vehicles. The DeepSORT algorithm is applied to track vehicle trajectories.
    \item \textbf{Trajectory Post-Processing Module:} The initial trajectories are converted using a perspective transformation matrix to map pixel coordinates to world coordinates. Additionally, the coordinate data undergoes filtering and noise reduction. The system computes motion state information for each detected object on a frame-by-frame basis, including position coordinates, longitudinal and lateral velocities, acceleration, and vehicle steering angles.
\end{itemize}
For urban village data, the trajectory dataset is generated through vehicle-mounted data collection, target detection, data synchronization, alignment, and scene segmentation.
\begin{itemize}
    \item \textbf{Data Collection Phase:} The data collection package is used to acquire data from six cameras, LiDAR point clouds, onboard IMU, and GPS sensors. Detected objects surrounding the vehicle are annotated.
    \item \textbf{Data Synchronization Phase:} Sensor data from different sources is temporally aligned.
    \item \textbf{Scene Segmentation Phase:} Interactions are classified based on interacting objects and actions.
    \item \textbf{Mapping and Trajectory Generation Phase:} The ego vehicle's position is used to construct a scene map, generating trajectory data that includes both the vehicle’s own trajectory and dynamic environmental information.
\end{itemize}
\begin{figure}
\centering
 \begin{minipage}{0.45\linewidth}
        \centering
 	\vspace{3.5pt}
    \centerline{\includegraphics[height=4.5cm, keepaspectratio]{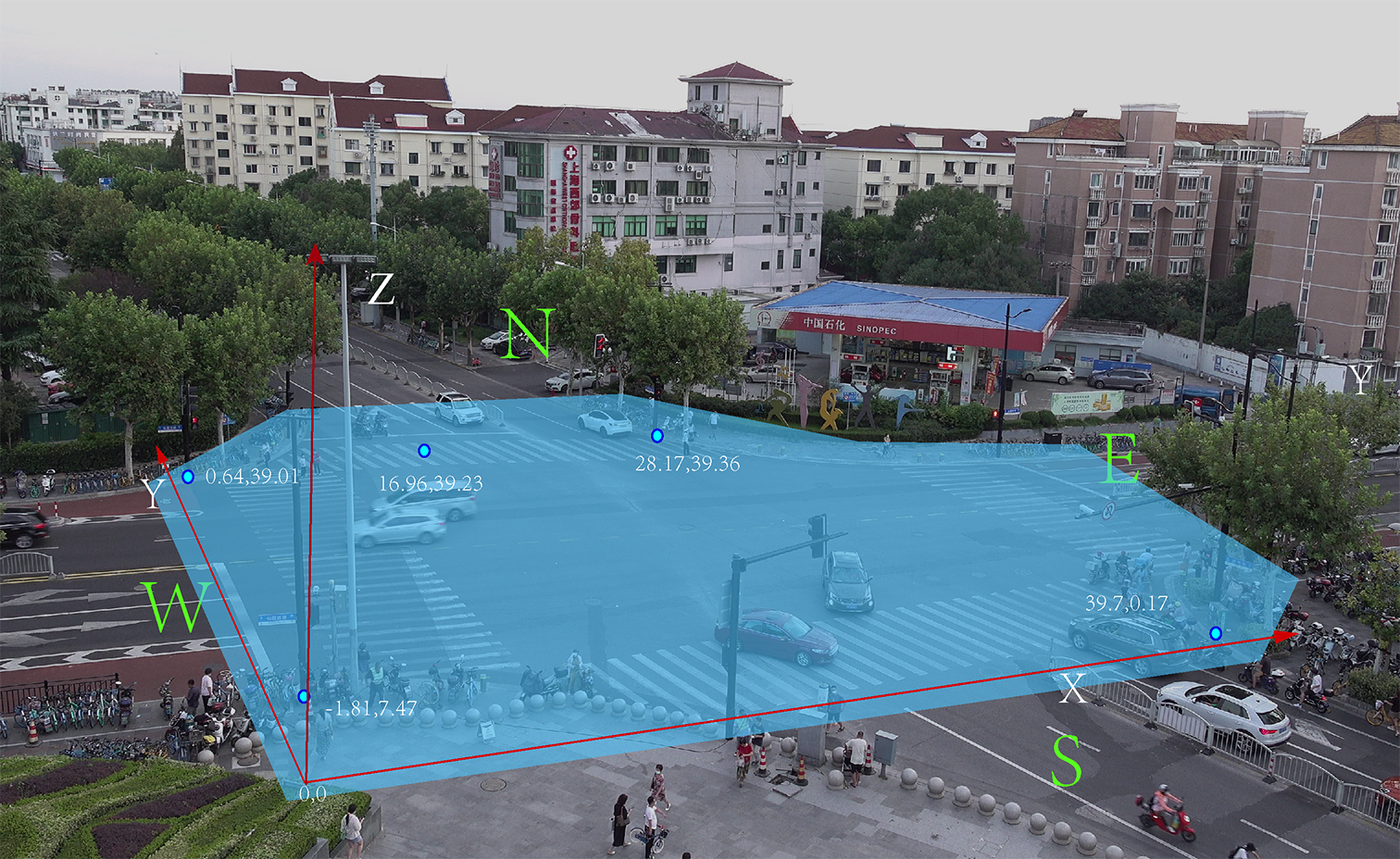}} 
        \centerline{a) Jianhe-Xianxia intersection}
 	\vspace{3.5pt}
 	\centerline{\includegraphics[height=4.5cm, keepaspectratio]{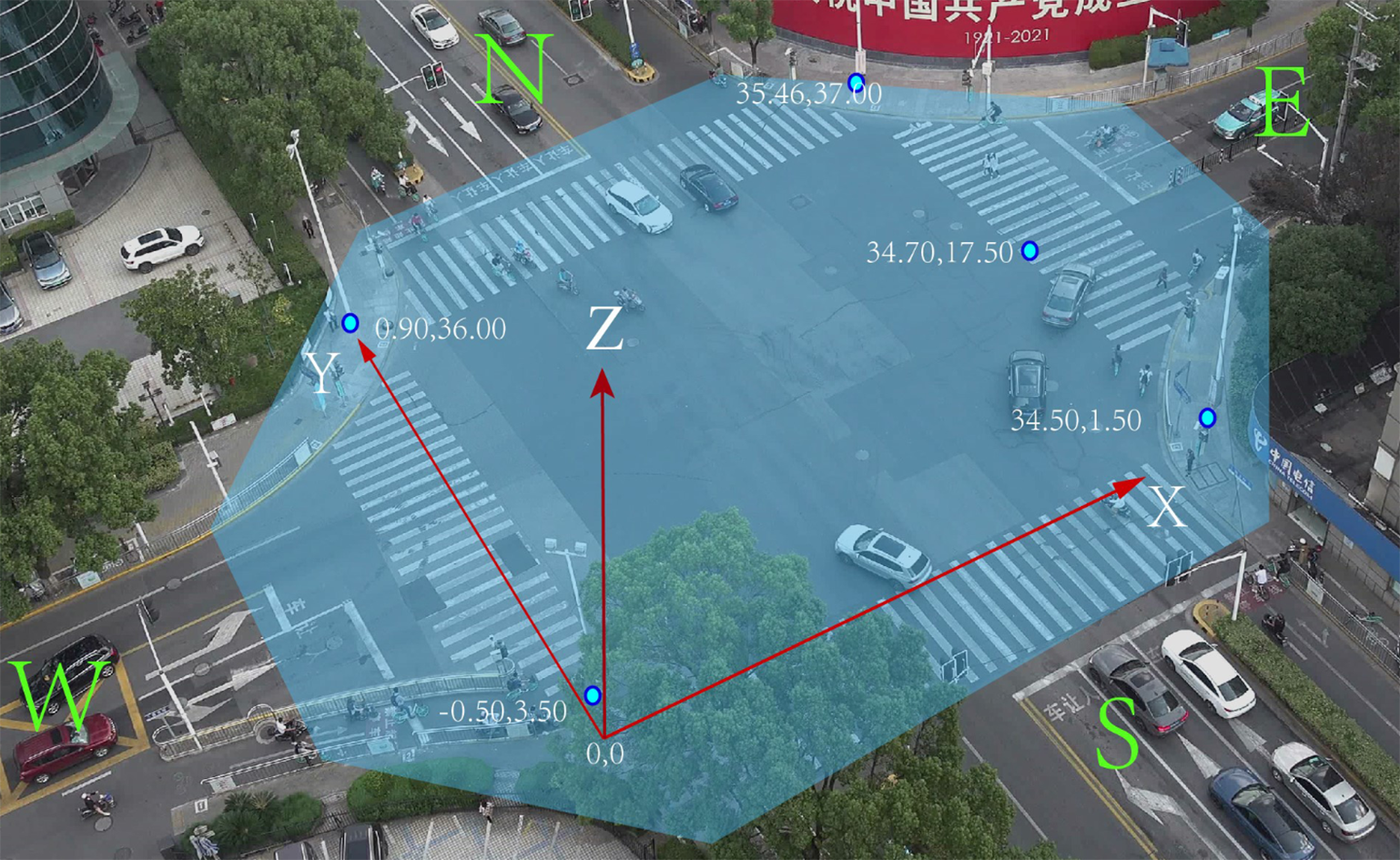}} 
 	\vspace{3.5pt}
 	\centerline{c) Moyu-Changji intersection}
        \vspace{3.5pt}
 \end{minipage}
 \begin{minipage}{0.45\linewidth}
        \centering
 	\vspace{3.5pt}
 	\centerline{\includegraphics[height=4.5cm, keepaspectratio]{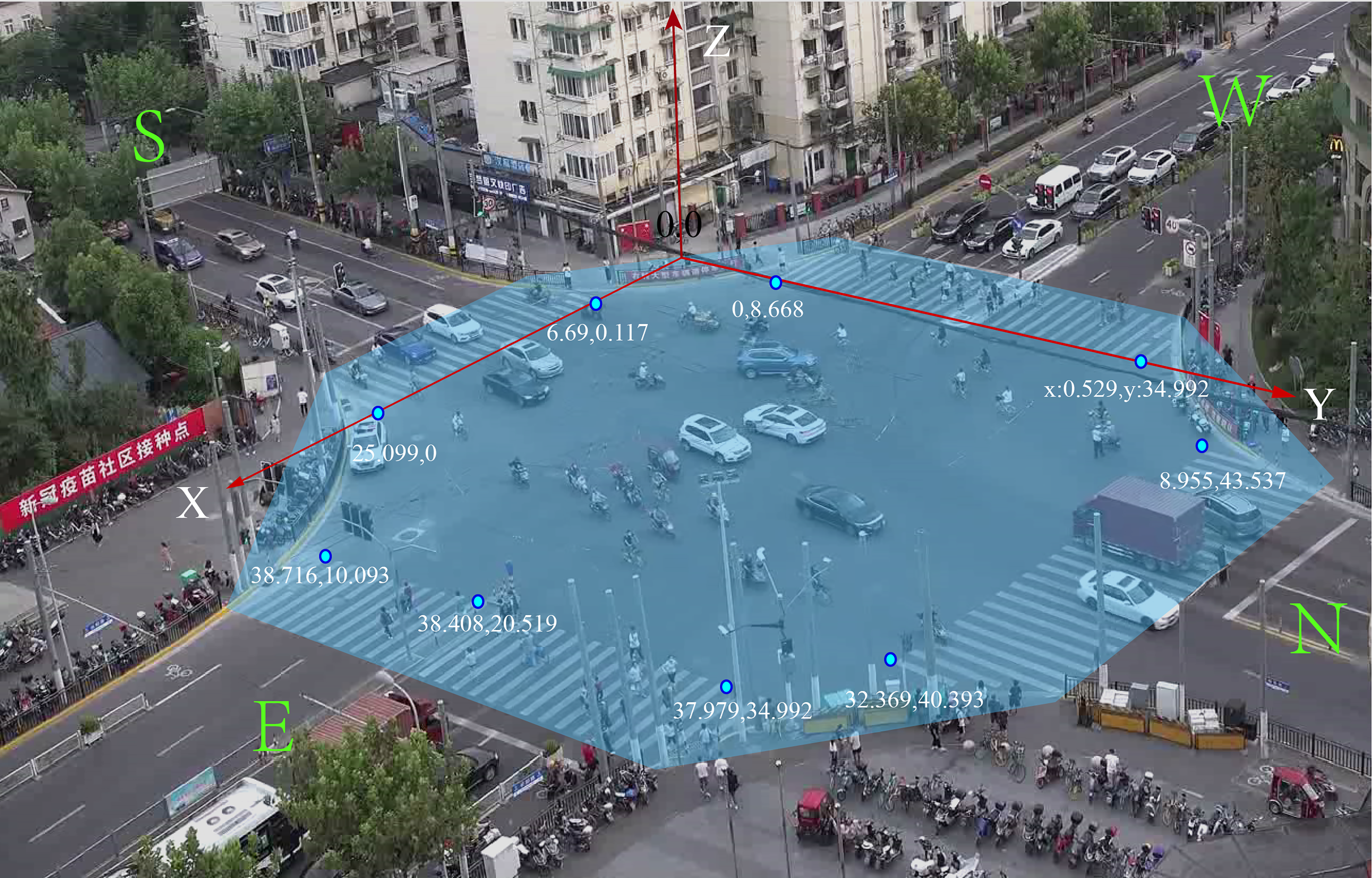}} 
        \centerline{b) Longchang-Jiyang intersection}
 	\vspace{3.5pt}
 	\centerline{\includegraphics[height=4.5cm, keepaspectratio]{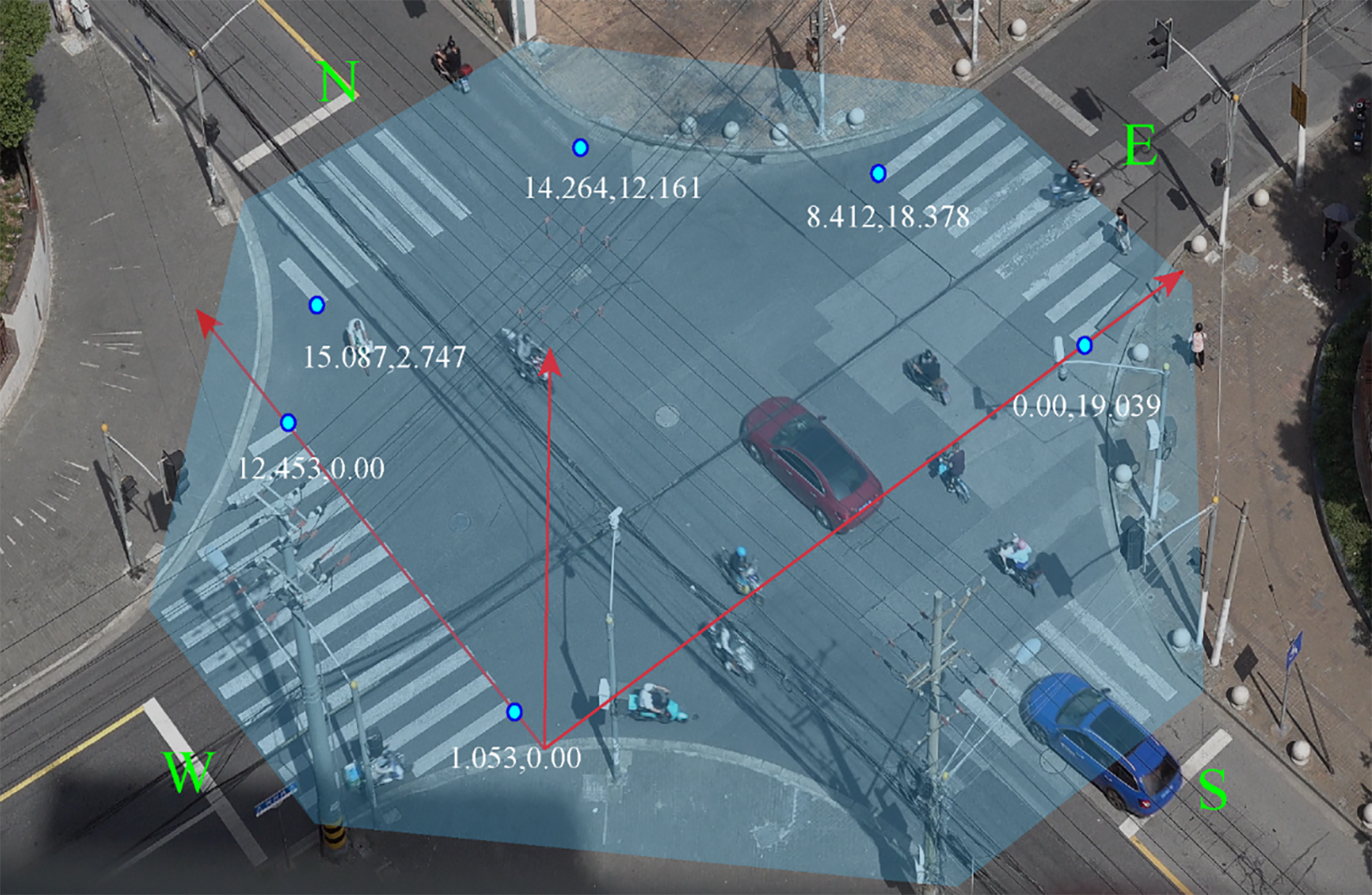}} 
 	\vspace{3.5pt}
 	\centerline{d) Ningwu-Hejian intersection}
  	\vspace{3.5pt}
 \end{minipage}
\caption{Coordinate System of the Intersection}
\label{fig:DISTRUBTION_YL}
\end{figure}

\subsection{Datasets and Formats}
Through the above processing, we obtained the initial data. To facilitate data application, we standardized the dataset by referring to the format of existing publicly available datasets. Specifically, this includes the standardization of vehicle trajectory data fields, alignment of intersection signal timing data, and high-precision map data, with the details as follows:

\begin{figure}
\centering
 \begin{minipage}{0.45\linewidth}
        \centering
 	\vspace{3.5pt}
    \centerline{\includegraphics[width=\textwidth]{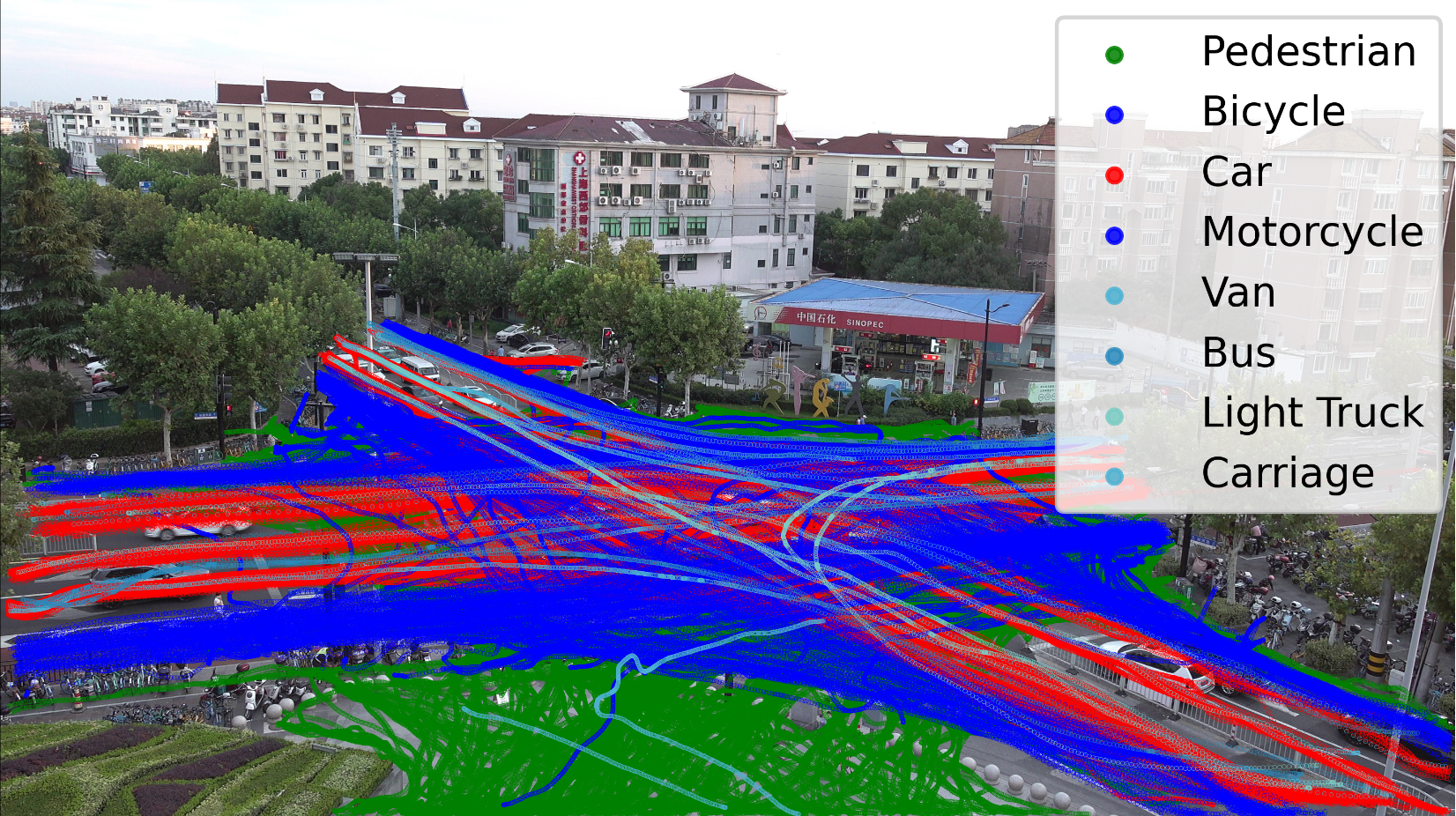}}
        \centerline{a) Trajectories at Jianhe-Xianxia intersection}
 	\vspace{3.5pt}
 	\centerline{\includegraphics[width=\textwidth]{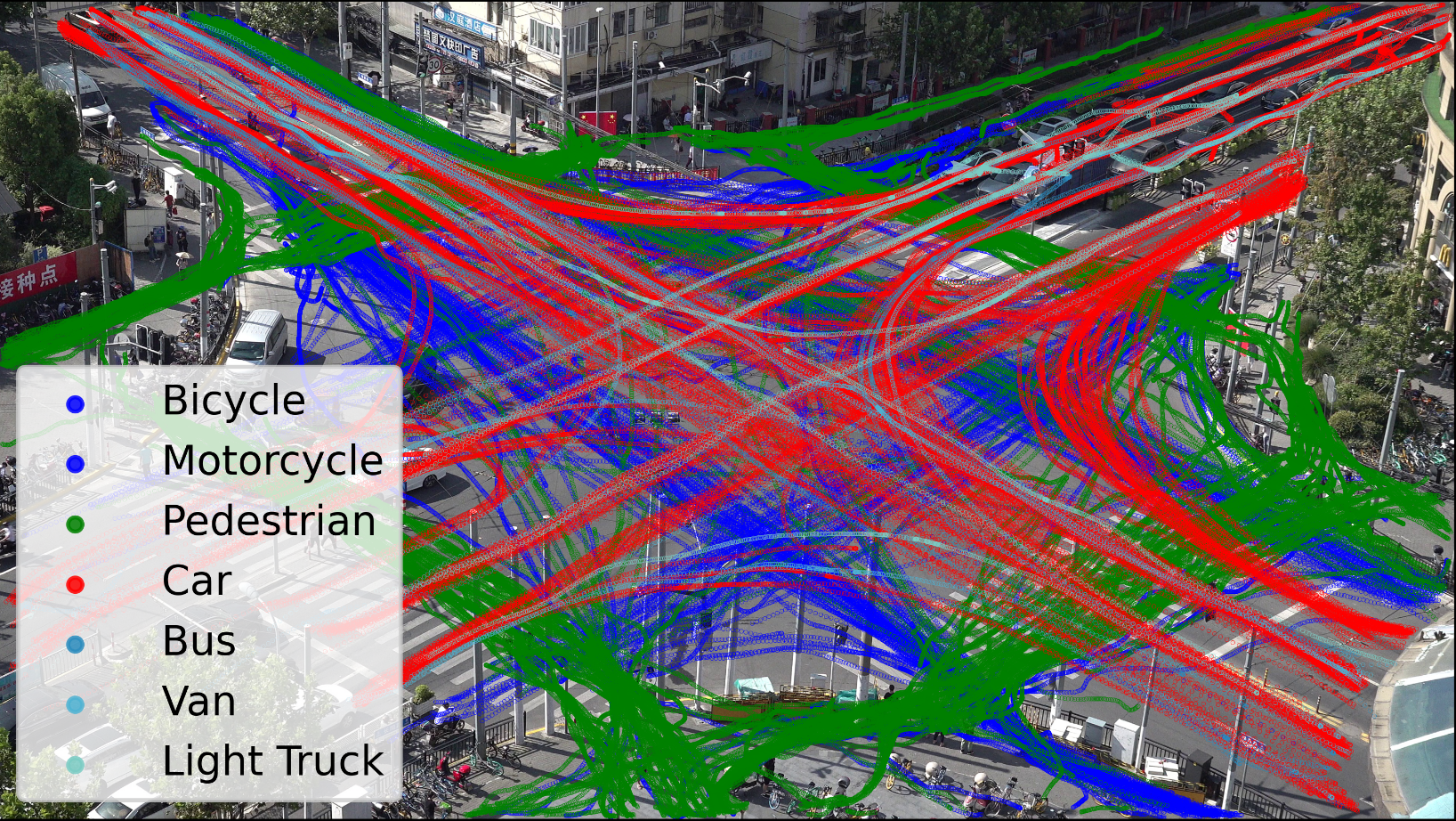}}
 	\vspace{3.5pt}
 	\centerline{c) Trajectories at Longchang-Jiyang intersection}
        \vspace{3.5pt}
 \end{minipage}
 \begin{minipage}{0.45\linewidth}
        \centering
 	\vspace{3.5pt}
 	\centerline{\includegraphics[width=\textwidth]{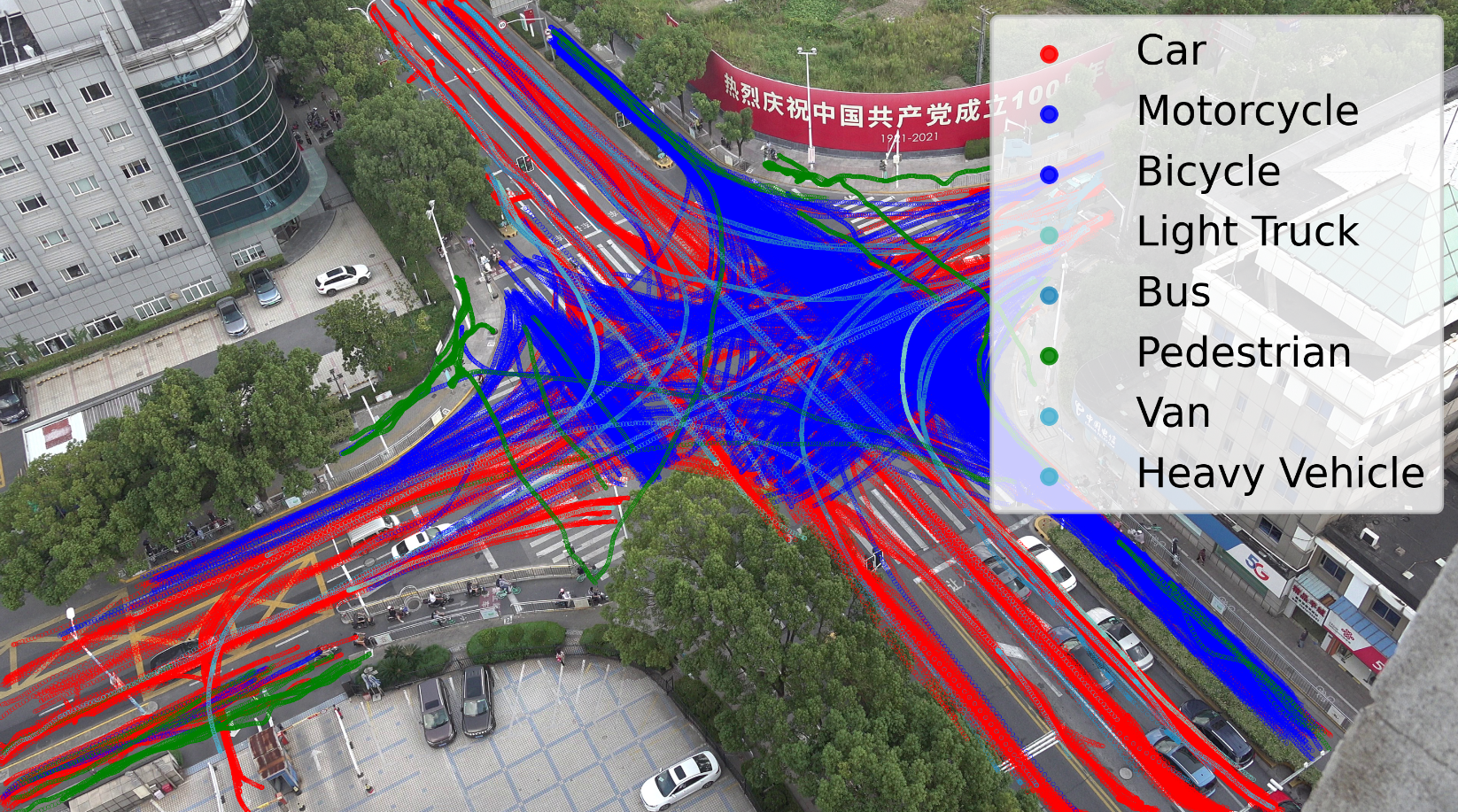}}
        \centerline{b) Trajectories at Moyu-Changji intersection}
 	\vspace{3.5pt}
 	\centerline{\includegraphics[width=\textwidth]{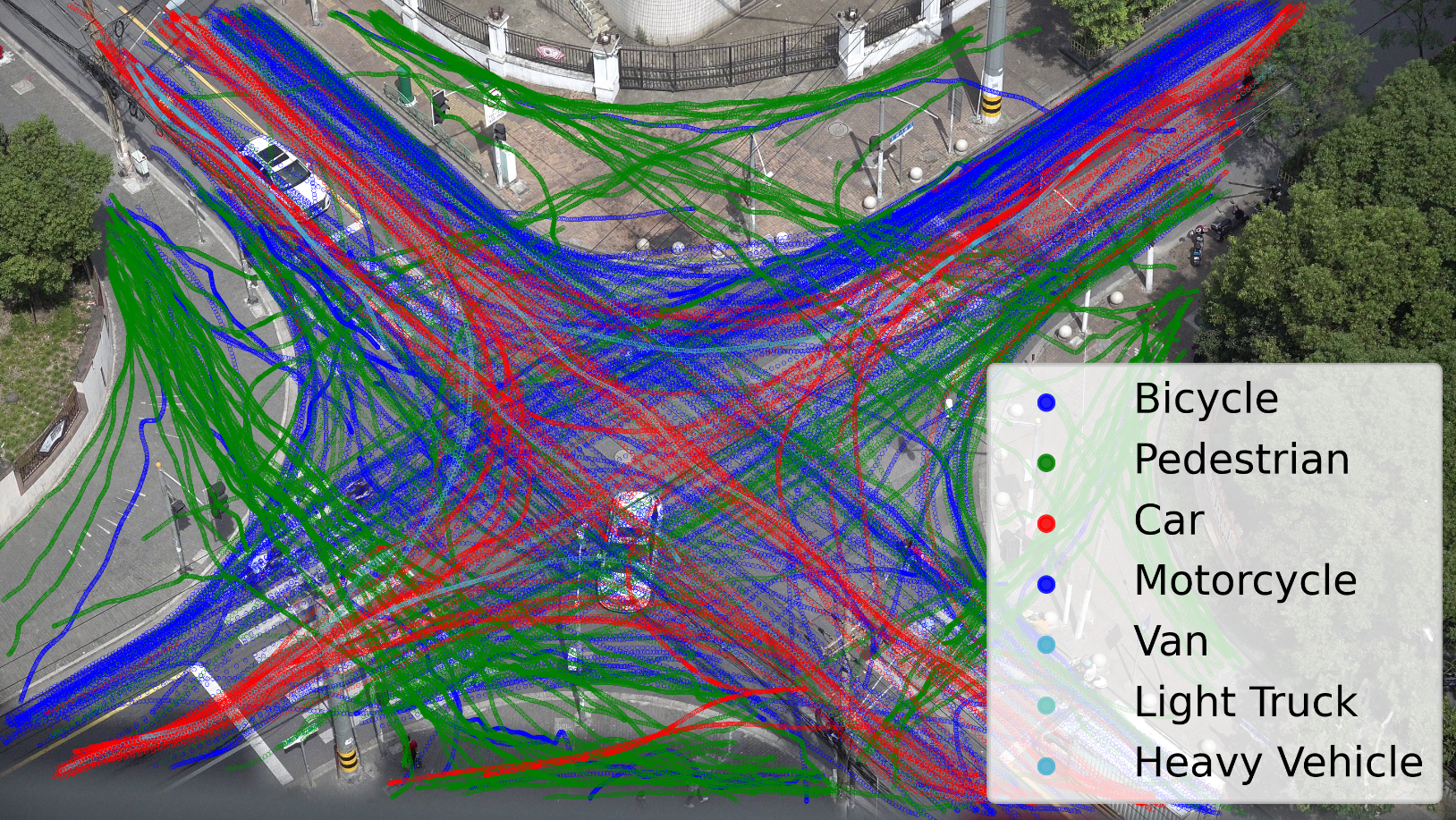}}
 	\vspace{3.5pt}
 	\centerline{d)Trajectories at Ningwu-Hejian intersection}
  	\vspace{3.5pt}
 \end{minipage}
\caption{The detailed information and trajectories at the intersections under study}
\label{fig:DISTRUBTION_YL2}
\end{figure}

\begin{itemize}
    \item \textbf{Trajectories and motion state:} The trajectory data in this VRU (Vulnerable Road Users) dataset captures the dynamic behavioral characteristics of traffic participants through multidimensional fields. Key fields include vehicle\_id (unique vehicle identifier) and vehicle\_type (covering nine types of traffic entities such as pedestrians, bicycles, and cars) to distinguish different objects by category. frame\_time (timestamp accurate to 0.04 seconds) combined with video\_id (video segment identifier in 10-minute intervals) provides a spatiotemporal reference framework. Spatial dynamic parameters include world\_x/world\_y (center coordinates of the vehicle bounding box) to locate entity positions, speed\_x/speed\_y (velocity components), acc\_x/acc\_y (acceleration components), and Jerk\_x/Jerk\_y (rate of acceleration change components) to describe changes in motion states, supplemented by Angle (angle between the movement direction and the x-axis) to characterize heading features. This high-precision spatiotemporal sequence data fully reconstructs the trajectory evolution of vehicles and pedestrians, supporting microscopic behavior analysis (e.g., lane-changing decisions, acceleration/deceleration patterns) and macroscopic traffic flow modeling. It is particularly useful for autonomous driving algorithm validation, traffic conflict risk assessment, and vulnerable road user safety research.

    \item \textbf{Traffic light states:} The trajectory data precisely records the dynamic behavior of intersection users and their correlation with signal control through multiple parameters. Core fields include inters\_nm (intersection name) and video\_id (video segment identifier) to define spatial and temporal data attribution. cycle\_id (signal cycle identifier) and phase (signal phase name), along with green/yellow/all\_red/red\_begin\_time (activation time of each signal light) and phase\_length (phase duration), fully describe the signal control timing logic. The linkage between trajectory dynamics (e.g., position, speed, acceleration) and signal states enables the quantification of vehicle/pedestrian start-stop characteristics under different signal phases, red-light running risks, and traffic efficiency. This “spatiotemporal-behavioral-control” three-dimensional coupling supports intersection conflict diagnosis, signal timing optimization, and autonomous driving cooperative decision-making research. It is particularly useful for revealing VRU behavior patterns in complex signal environments and developing safety enhancement strategies. The signal control scheme can be associated with the trajectory dataset through intersection names, video identifiers, and signal phase start/end times.
    \item \textbf{High-Definition (HD) Map:} In addition to trajectory recording files, this dataset provides a high-definition (HD) intersection map compliant with the Lanelet2 format, whose origin is aligned with the ground coordinate system origin of the trajectory data. The map contains semantic information such as lane topology and traffic regulations, supporting multi-modal perception and path planning for autonomous driving. By integrating spatiotemporal behavioral parameters (position, velocity, signal phase) from the trajectory data with the lane-level structure of the Lanelet2 map, dynamic interactions between vehicles/pedestrians under real-world signal control can be precisely analyzed (e.g., lane-level trajectory prediction, signal compliance verification). Referencing the OSM map conversion methodology from the SinD case, this data architecture is extendable to other scenarios, offering a multi-source fusion benchmark for autonomous driving simulation, vehicle-infrastructure cooperative decision-making, and mixed traffic flow modeling.
\end{itemize}

\section{Statistics and Evaluation}
Statistical analysis of existing publicly available trajectory datasets reveals significant differences in information coverage. Taking typical datasets as examples, although INTERACTION is the largest (containing approximately 18,000 trajectories), only 20.4\% of its data pertains to signalized intersections, and it lacks both traffic signal status and pedestrian/bicycle data, limiting behavior correlation analysis. The inD dataset, while providing complete traffic signal phase information, is restricted in terms of scene diversity and participant types, primarily focusing on motor vehicles. In contrast, the Onsite VRU dataset proposed in this study covers 100\% of trajectories at signalized intersections (including 17,429 trajectories) and integrates high-definition maps, signal timing details (phase activation/deactivation times, cycle numbers), and refined motion parameters (speed, acceleration, jerk, and heading angle) for nine categories of traffic participants, including pedestrians and non-motorized vehicles. The dataset achieves a labeling completeness rate of 98.7\%. Statistical findings indicate that existing datasets commonly suffer from missing signal-behavior coupling data (86\% lack traffic signal information) and a low proportion of vulnerable road users (VRUs) (averaging less than 50\%). By incorporating multi-dimensional data fusion and diverse scene designs, Onsite VRU provides a more comprehensive benchmark for mixed traffic flow modeling and autonomous driving decision-making.

\subsection{Scale and Number of Categories}

In our established \textbf{OnsiteVUR dataset}, the proportion of VRUs (Vulnerable Road Users) exceeds 50\%, significantly. This high proportion endows the dataset with unique advantages in revealing the behavioral characteristics and adaptive performance of VRUs. Compared to traditional traffic datasets, the \textbf{OnsiteVUR} dataset features a higher density of VRUs, enabling a more comprehensive capture of the dynamic behaviors of pedestrians and non-motorized users in complex traffic environments. This high-density distribution of VRUs provides rich data support for studying their interaction patterns with motor vehicles, risk assessment, and safety strategies. By analyzing this dataset, we can gain deeper insights into the behavioral patterns of VRUs in real-world scenarios, such as their route choices, speed variations, and responses to traffic signals. Furthermore, the high proportion of VRU data also offers critical references for developing intelligent transportation systems and autonomous driving technologies, particularly in enhancing VRU safety and adaptive performance. As shown in the figure, the proportion of VRUs in the dataset is significantly higher than that of other traffic participants, further validating the unique value and application potential of the \textbf{OnsiteVUR} dataset in the field of VRU research.

\begin{figure}[htbp]
  \centering
  \begin{subfigure}[b]{0.45\textwidth}
    \includegraphics[width=\textwidth]{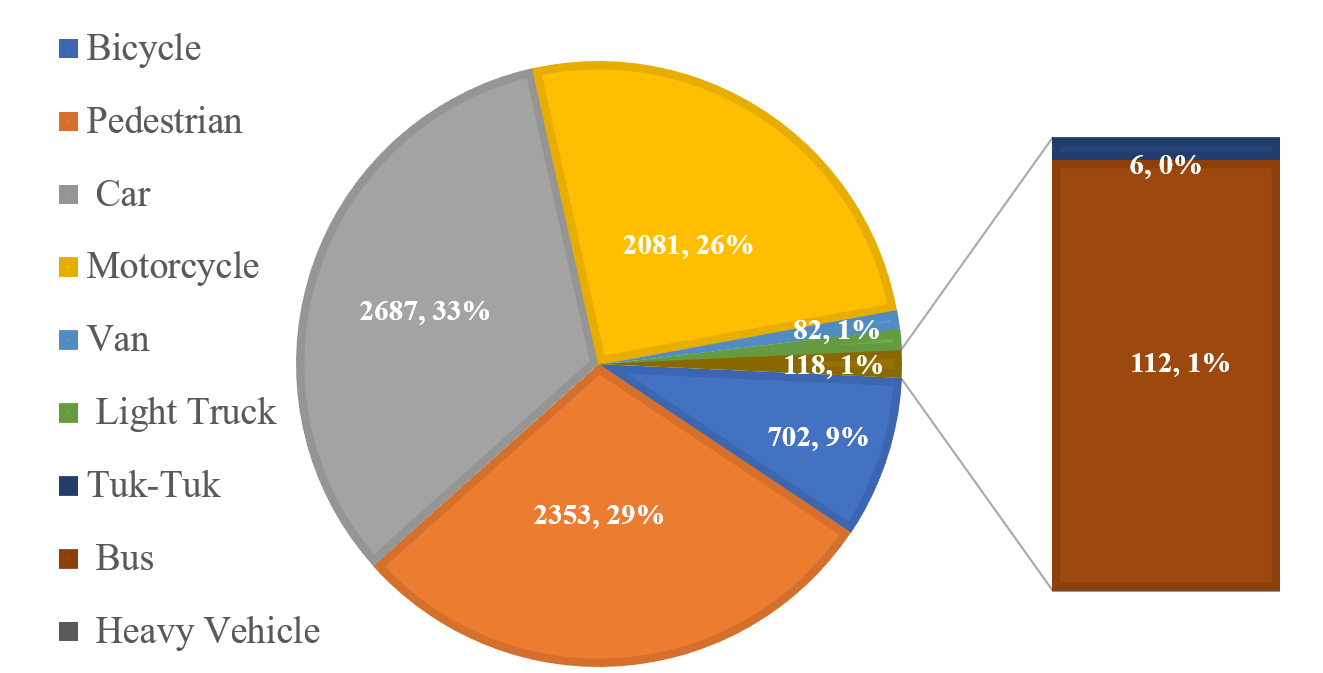}
    \caption{LC}
    \label{fig:image_a1}
  \end{subfigure}
  \hfill 
  \vspace{3.5pt}
  \begin{subfigure}[b]{0.45\textwidth}
    \includegraphics[width=\textwidth]{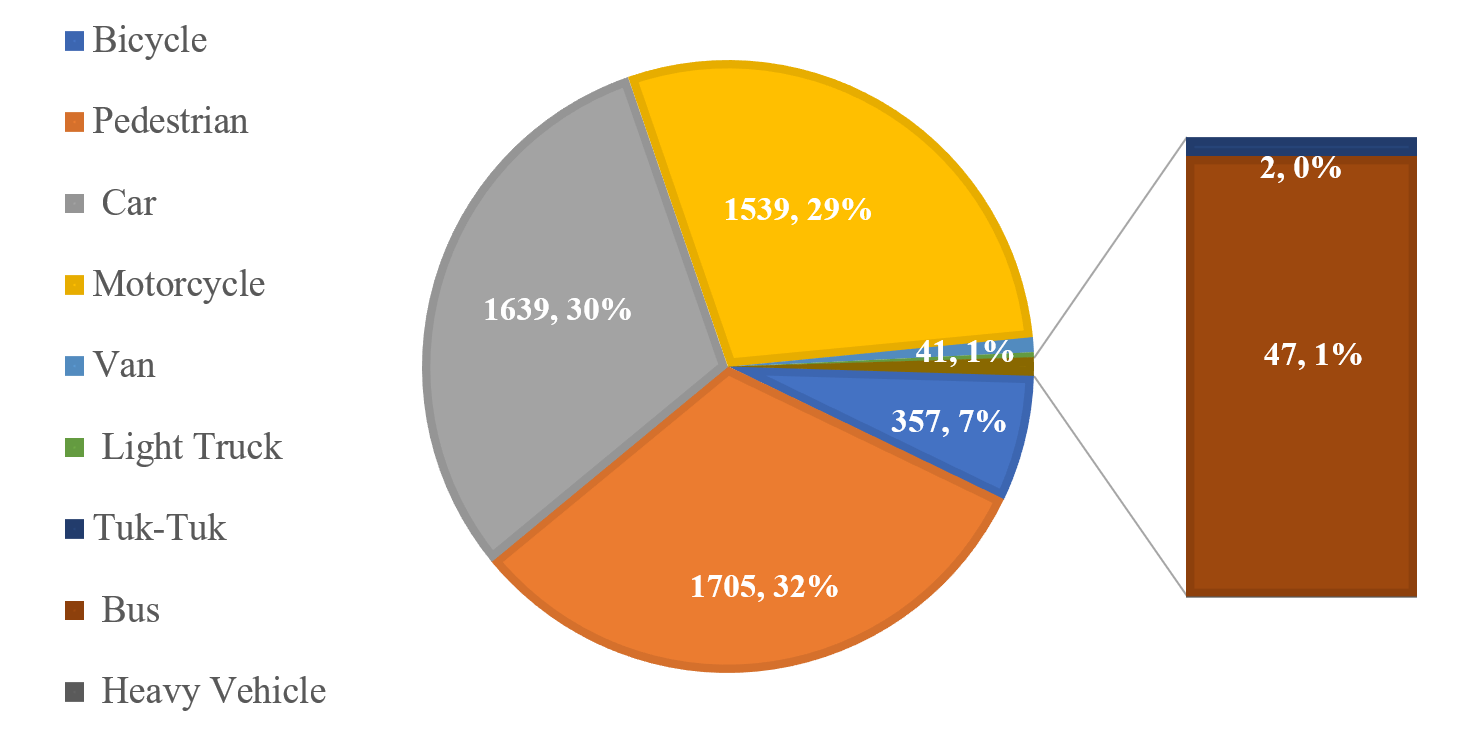}
    \caption{JH}
    \label{fig:image_b1}
  \end{subfigure}

  \begin{subfigure}[b]{0.45\textwidth}
    \includegraphics[width=\textwidth]{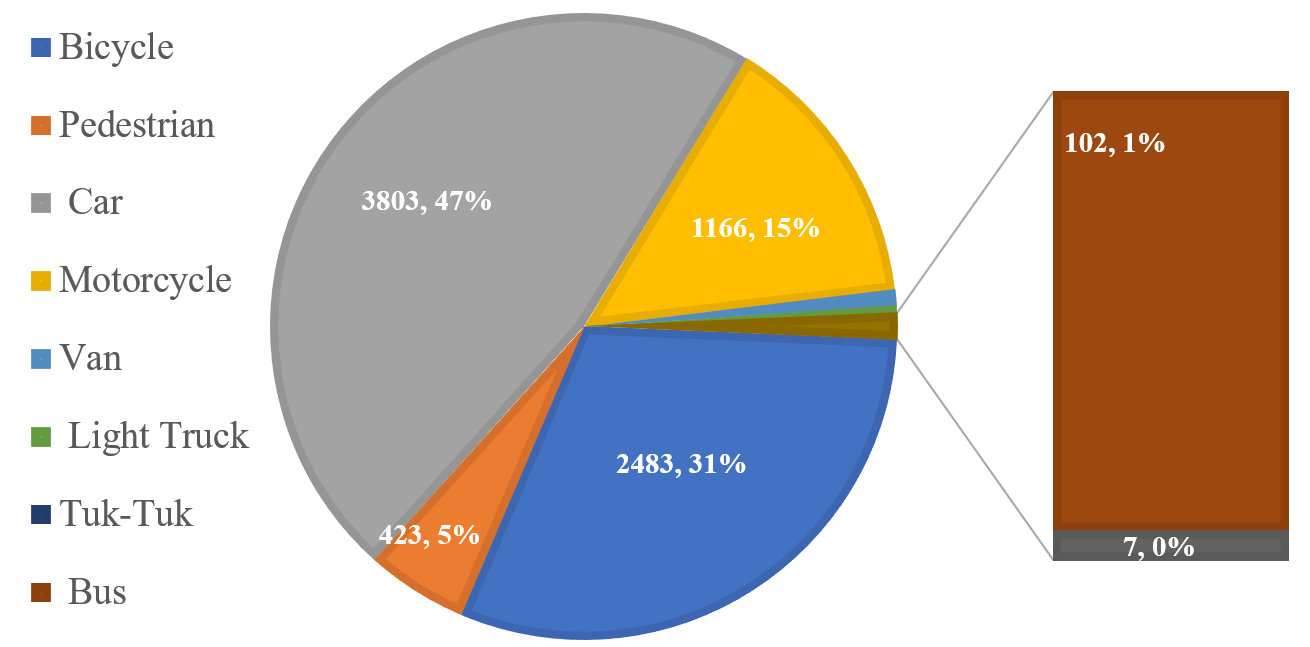}
    \caption{MY}
    \label{fig:image_c1}
  \end{subfigure}
  \hfill
  \vspace{3.5pt}
  \begin{subfigure}[b]{0.45\textwidth}
    \includegraphics[width=\textwidth]{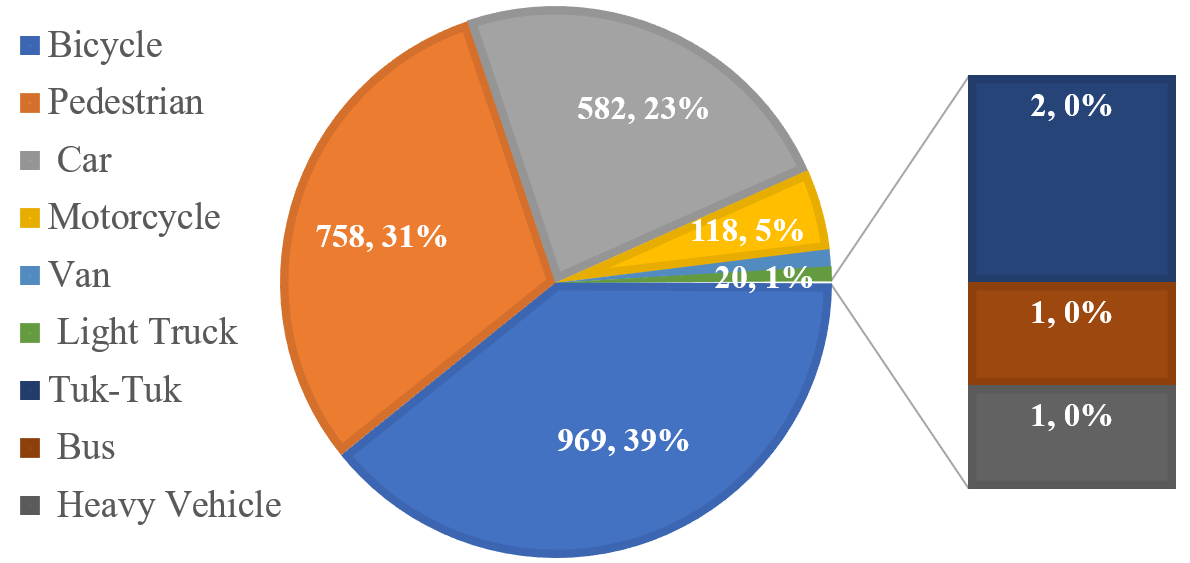}
    \caption{NW}
    \label{fig:image_a2}
  \end{subfigure}

  \begin{subfigure}[b]{0.45\textwidth}
    \includegraphics[height=3.8cm]{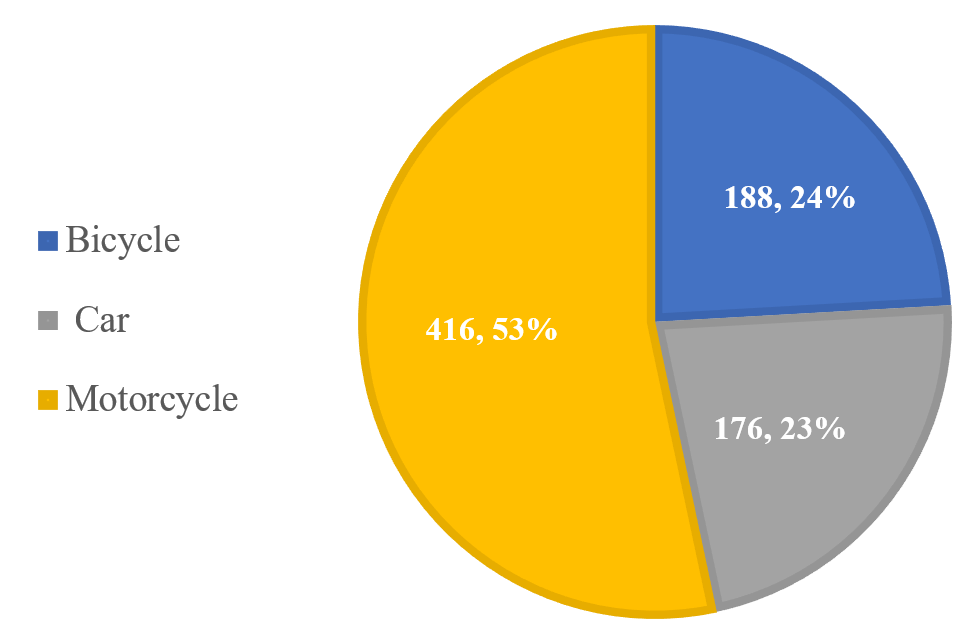} 
    \caption{AY}
    \label{fig:image_b2}
  \end{subfigure}
  \hfill
  \vspace{3.5pt}
  \begin{subfigure}[b]{0.45\textwidth}
    \includegraphics[height=3.8cm]{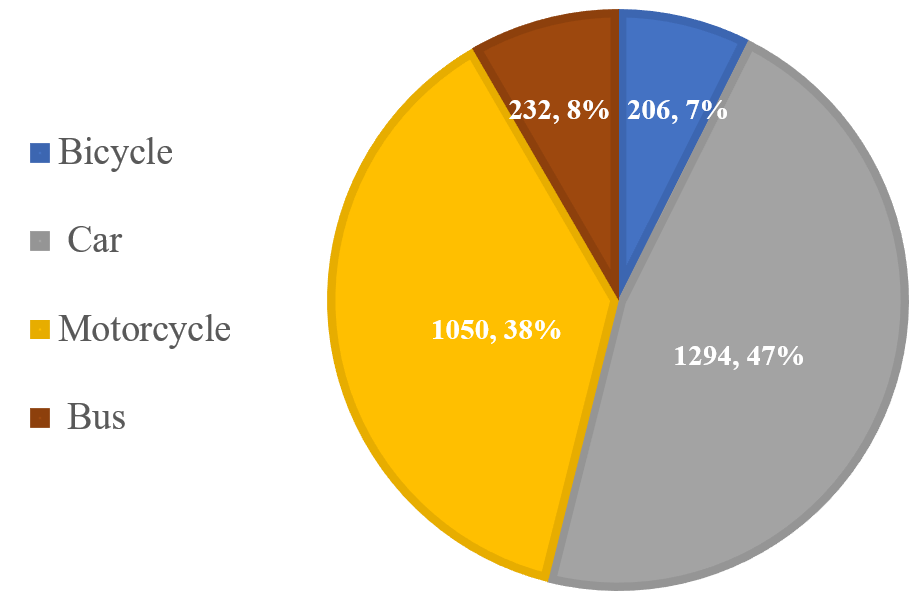} 
    \caption{CY}
    \label{fig:image_c2}
  \end{subfigure}

  \caption{Proportion of Each Type of Vehicle in the Dataset}
  \label{fig:multi_images}
\end{figure}

\subsection{VRU Conflict Distribution}

\begin{figure}
\centering
 \begin{minipage}{0.48\linewidth}
        \centering
 	\vspace{3.5pt}
    \centerline{\includegraphics[height=4.5cm, keepaspectratio]{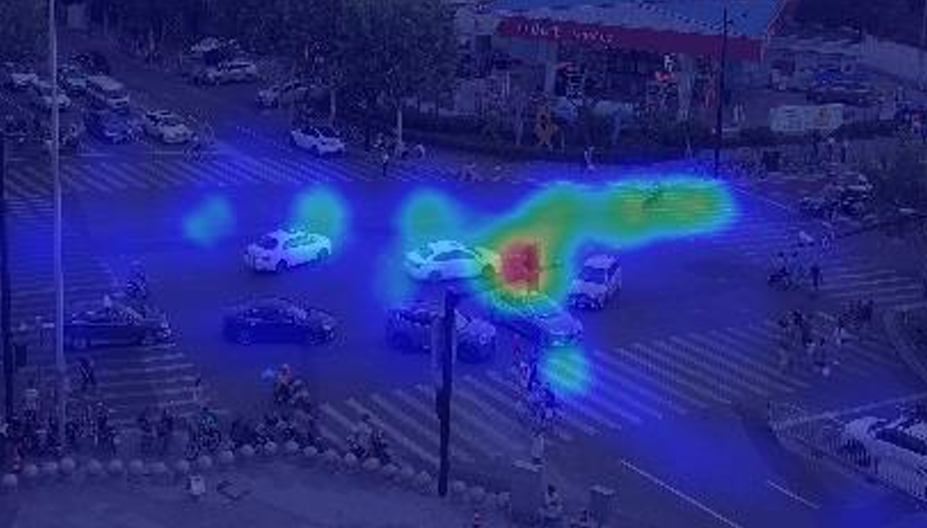}} 
        \centerline{a) Jianhe-Xianxia intersection}
 	\vspace{3.5pt}
 	\centerline{\includegraphics[height=5cm, keepaspectratio]{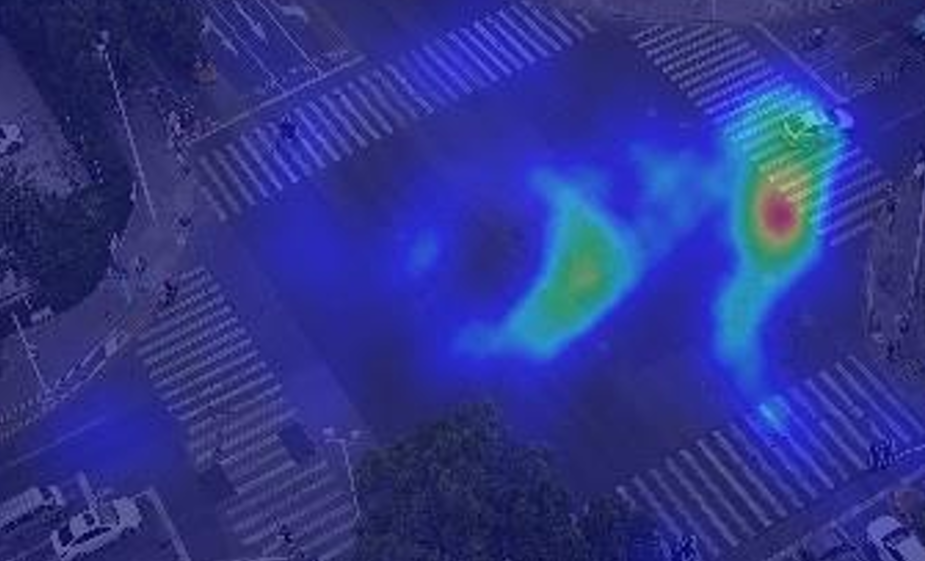}} 
 	\vspace{3.5pt}
 	\centerline{c) Moyu-Changji intersection}
        \vspace{3.5pt}
 \end{minipage}
 \begin{minipage}{0.475\linewidth}
        \centering
 	\vspace{3.5pt}
 	\centerline{\includegraphics[height=4.5cm, keepaspectratio]{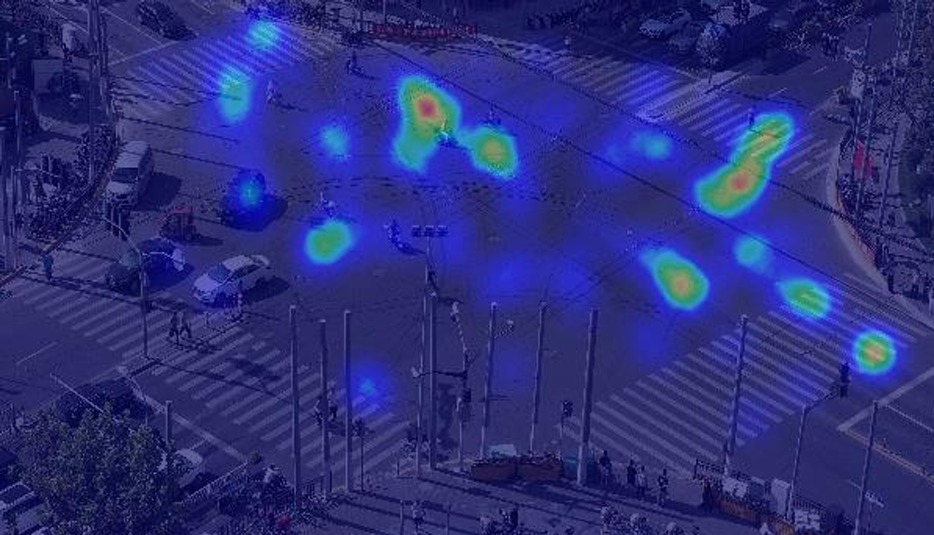}}
        \centerline{b) Longchang-Jiyang intersection}
 	\vspace{3.5pt}
 	\centerline{\includegraphics[height=5cm, keepaspectratio]{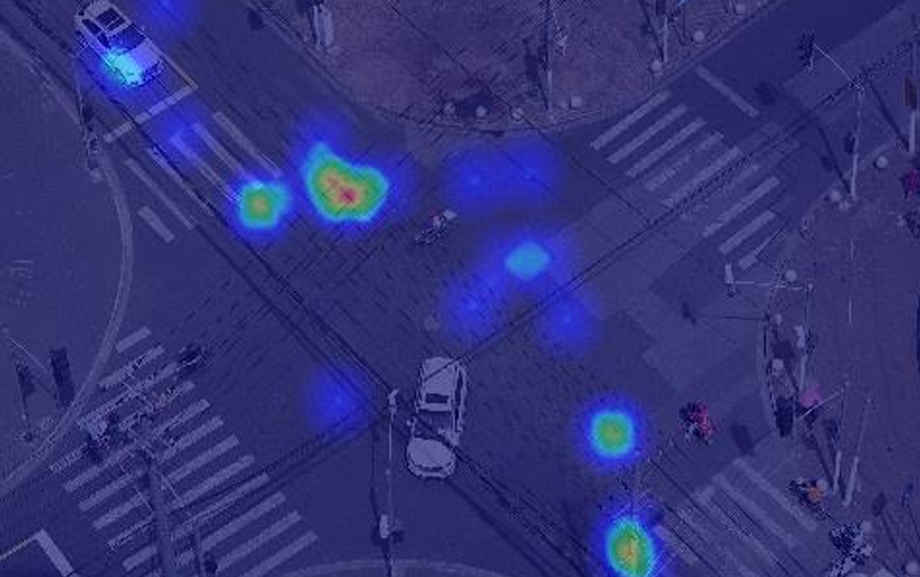}} 
 	\vspace{3.5pt}
 	\centerline{d) Ningwu-Hejian intersection}
  	\vspace{3.5pt}
 \end{minipage}
\caption{Distribution of Conflicts Between Motor Vehicles and Non-Motorized Vehicles at Intersections}
\label{fig:conflict}
\end{figure}

Based on the aforementioned trajectory data, an analysis of high-risk motor vehicle and vulnerable road user (VRU) interactions at intersections was conducted. As shown in Figure Figure~\ref{fig:conflict}, VRU conflicts at the Moyu Road–Changji East Road intersection are primarily concentrated in the east entrance left-turn, south entrance right-turn, and west entrance left-turn areas, which align well with actual traffic conditions. This intersection employs a signal control scheme that allows left-turning, through, and right-turning motor vehicles and VRUs to proceed simultaneously in the east-west direction, leading to crossing conflicts between eastbound left-turning motor vehicles and westbound through-moving VRUs. Additionally, westbound left-turning VRUs experience merging conflicts with eastbound right-turning motor vehicles, as well as crossing conflicts with eastbound through-moving motor vehicles. The most severe conflicts at this intersection occur between eastbound left-turning motor vehicles and westbound through-moving VRUs, as well as between westbound left-turning VRUs and eastbound right-turning motor vehicles.

As shown in Figure~\ref{fig:conflict}, general VRU conflicts at the Changyang Road–Longchang Road intersection are mainly concentrated in front of the north entrance through lane, the south entrance left-turn lane, and the merging area of the west exit lane. This intersection adopts a mixed-phase signal control scheme, allowing left-turning, through, and right-turning movements in the north-south direction simultaneously. As a result, left-turning VRUs must weave through both same-direction and opposing through traffic, leading to high-density crossing conflicts, particularly between south entrance left-turning motor vehicles and northbound VRUs, as well as between north entrance right-turning motor vehicles and northbound VRUs. Additionally, severe conflicts at this intersection are mainly concentrated between north entrance VRUs and north-south through-moving motor vehicles, as well as between south entrance left-turning motor vehicles and northbound VRUs.

\subsection{Comparison and Discussion}

To highlight the differences, this study compares the OnSiteVRU dataset with two recently developed, similar datasets (SIND and INTERACTION). The results are presented in Table I and Figure 7. While all three datasets provide high-definition maps (HD maps), which are crucial for analyzing and predicting the behavior of traffic participants, they differ significantly in terms of scene coverage, dataset scale, traffic signal state information, and diversity of traffic participants.

First, the INTERACTION dataset has the most extensive scene coverage and the largest scale, containing 18,471 trajectories of traffic participants at intersections. Among them, 14,867 trajectories are from unsignalized intersections, while 3,775 trajectories are from signalized intersections. However, it lacks signal state information throughout the dataset [1]. Additionally, based on the provided dataset snapshots, signalized intersections should include a large number of pedestrians, yet the corresponding files do not provide any pedestrian or bicycle data, raising concerns about data completeness. In contrast, the SIND dataset includes signal state information but has relatively limited scene coverage, focusing mainly on urban intersections. Moreover, it primarily features motor vehicles, lacking sufficient data on non-motorized vehicles and pedestrian behavior.

The OnSiteVRU dataset demonstrates significant advantages in both scene diversity and traffic participant coverage. It includes multiple intersection and roadway scenarios, covering a wide range of traffic participants, including motor vehicles, electric bicycles, and human-powered bicycles. The intersection scenarios include Xianxia Road-Jianhe Road, Changyang Road-Longchang Road, Moyu Road-Changji East Road, and Hejian Road-Ningwu Road, with a total of approximately 17,429 trajectories. These trajectories comprehensively record signal state information and the dynamic behavior of traffic participants. The roadway segments include Caoyang Road (a lane-separated segment with 50 minutes of data and 1,256 trajectories) and Anyuan Road (a bidirectional two-lane roadway without separation, with 1 hour and 40 minutes of data), which reflect the traffic flow characteristics under different road designs. Furthermore, the OnSiteVRU dataset also covers urban village scenarios (collected from Jiading District in Shanghai and Pingjiang District in Suzhou), capturing high-density mixed traffic behavior in unstructured road environments. This fills a gap in existing datasets regarding complex scene coverage.

Overall, the OnSiteVRU dataset outperforms the INTERACTION and SIND datasets in terms of signal state completeness, diversity of traffic participants, and scene coverage. It not only provides high-definition maps and detailed signal state information but also includes diverse scenarios ranging from urban intersections to unstructured roads. This enables a more comprehensive study of vulnerable road user (VRU) behavior in mixed traffic environments, offering high-quality data support for autonomous driving algorithm optimization and traffic safety management.

\section{Application of Autonomous Driving Virtual Testing Platform}

As road traffic environments become increasingly complex, protecting Vulnerable Road Users (VRUs) has become a critical issue in traffic safety research. VRUs typically refer to groups that are more vulnerable in traffic interactions, including non-motorized vehicles and pedestrians. A deep understanding and accurate prediction of the behavior patterns of these groups are crucial for improving traffic safety. In complex road environments, VRUs face numerous potential risks, such as traffic rule violations, high-speed driving, and other risk behaviors, which can result in severe injuries or traffic accidents. Therefore, accurately identifying and predicting the behavior trajectories of VRUs and conducting scientific assessments has become a core task in ensuring road safety.

\subsection{Challenge Format}
This challenge adopts an online release, offline deployment, and prediction approach, with a real-time leaderboard update, aiming to comprehensively assess the effectiveness of trajectory prediction algorithms in ensuring the safety of Vulnerable Road Users (VRUs) in a fair and transparent manner. The specific competition process is as follows:

\begin{enumerate}
    \item \textbf{Dataset and Evaluation Metric Release}: The organizers will first release a standardized dataset along with the clear definition of evaluation metrics. The dataset contains real VRU behavior trajectories, covering various complex traffic scenarios and involving trajectories of pedestrians, non-motorized vehicles, and other vulnerable road users. Additionally, the organizers will release a unified set of evaluation metrics for participants to measure the performance of different algorithms in the trajectory prediction task. It is important to note that multiple datasets of varying completeness will be released, with the aim of evaluating the prediction performance of participants on different datasets.

    \item \textbf{Download Data and Locally Deploy Algorithm for Testing}: After downloading the dataset, participants are required to use their locally deployed trajectory prediction algorithms to predict missing trajectories in the test dataset. Participants must complete the prediction and generate result files within the specified timeframe, then submit the predicted data files to the online platform for evaluation by the organizers.~\url{https://www.kaggle.com/datasets/zcyan2/onsitevru-trajectory-prediction-dataset}.

    \item \textbf{Organizer Evaluation and Leaderboard Update}: The organizers will strictly evaluate the submitted prediction results based on the predefined evaluation metrics. The evaluation process will score according to each standard and update the competition leaderboard in real time based on the scores. Participants can view their performance and relative ranking on each evaluation metric through the leaderboard, enabling them to comprehensively understand the strengths and areas for improvement of their algorithms.
\end{enumerate}

\subsubsection{Trajectory Format}

The dataset includes the training set (\texttt{train\_data\_x.npy}), training labels (\texttt{train\_data\_y.npy}), and test set (\texttt{test\_data\_x.npy}), all saved in \textbf{.npy} format. The data format is a 4-dimensional matrix with the following dimensions: [Scene ID, Time Step, Participant ID, Feature Value]. Each feature value consists of 10 columns.

\subsubsection{Evaluation Metrics}
To calculate the overall score of the model, this competition combines multiple evaluation metrics (\textit{minADE}, \textit{minFDE}, \textit{MissRate}, and \textit{mAP}) and computes the total score by summing the weighted results. The specific methodology is as follows:
\begin{equation}
    \text{score} = 0.2 \cdot S_{\text{minADE}} + 0.2 \cdot S_{\text{minFDE}} + 0.3 \cdot S_{\text{MR}} + 0.2 \cdot S_{\text{w-mAP}}
\end{equation}

\begin{itemize}
    \item \textbf{Minimum Average Displacement Error (min ADE)}
    \begin{equation}
    \text{minADE} = \min\left(\frac{1}{T}\sum_{t=1}^{T}\sqrt{(x_{k,t} - x_{\text{gt},t})^2 + (y_{k,t} - y_{\text{gt},t})^2}\right) \quad k \in \{1,\ldots,K\}
    \end{equation}
    \item \textbf{Minimum Final Displacement Error (minFDE)}
    \begin{equation}
     \text{minFDE} = \min\left(\sqrt{(x_{k,T} - x_{\text{gt},T})^2 + (y_{k,T} - y_{\text{gt},T})^2}\right) \quad k \in \{1,\ldots,K\}
    \end{equation}
    \item  \textbf{Weighted Mean Average Precision (w-mAP)}
    \begin{equation}
    \text{w-mAP} = \sum_{\tau \in T} w(\tau) \cdot \text{AP}(\tau)
    \end{equation}
\end{itemize}

where \(\tau \in \{0.5, 1, 1.5\}\) with weights \([0.6, 0.3, 0.1]\).

\section{Conclusion}
This paper introduces the construction process, main content, and application cases of the OnSiteVRU dataset in traffic research. The OnSiteVRU dataset collects multimodal data from typical Chinese traffic scenarios (such as intersections, road segments, and urban villages) through high-precision sensors (e.g., LiDAR, cameras) and multi-source data fusion technology. It covers trajectory information of various traffic participants, including motor vehicles, electric bicycles, and human-powered bicycles. The dataset includes multiple intersection scenarios, such as Xianxia Road-Jianhe Road, Changyang Road-Longchang Road, Moyu Road-Changji East Road, and Hejian Road-Ningwu Road, as well as road segment scenarios like Caoyang Road and Anyuan Road, totaling approximately 17,429 trajectories. Additionally, the dataset encompasses unstructured road scenarios like urban villages, capturing high-density mixed traffic behaviors. The OnSiteVRU dataset has been applied in traffic flow modeling, trajectory prediction, and autonomous driving virtual testing, providing essential data support for VRU behavior research in mixed traffic environments.

Compared to traditional open-source datasets, the OnSiteVRU dataset has significant advantages. First, it features higher VRU density and ratio, with broader scenario coverage, enabling a more comprehensive reflection of VRU behavioral characteristics in mixed traffic flows. Second, the data precision reaches 0.04 seconds, offering rich microscopic behavioral information and laying a solid foundation for detailed traffic flow modeling and risk analysis. Moreover, the dataset combines natural driving data from a top-down perspective and real-time dynamic detection data from an onboard perspective, meeting diverse needs such as traffic flow modeling, trajectory prediction, and autonomous driving virtual testing. Additionally, the dataset includes environmental information such as traffic lights, obstacles, and real-time maps, enabling a more comprehensive reconstruction of interaction events and providing high-value data for behavior research in complex scenarios.

However, the OnSiteVRU dataset still has some limitations. For example, the collection features of intersection and road segment map data are relatively homogeneous. The team is currently measuring higher-precision map information and plans to develop more scenarios, such as T-junctions, to further enhance the dataset's diversity and practicality. Furthermore, the dataset's coverage of extreme conditions (e.g., adverse weather, nighttime environments) remains limited, and future efforts will focus on expanding scenario types and data dimensions.

In summary, the OnSiteVRU dataset provides significant support for mixed traffic flow research and the development of autonomous driving technologies through high-quality and diverse data collection and processing. In the future, with further optimization of multi-source sensing technologies and continuous improvement of the data ecosystem, the OnSiteVRU dataset will continue to expand its scenario coverage and data dimensions, offering stronger support for the localized deployment of global autonomous driving systems and traffic safety research.

\bibliography{samplett}
\end{document}